# Is Trump's Vocabulary Poor?
# Vocabulary Richness Across Texts of Different Lenghts

Dominique Labbé [1], Cyril Labbé [2] (0000-0002-1189-2297), Jacques Savoy [3*] (0000-0002-4486-0067)

[1] Science Po, Université Grenoble Alpes
[2] Laboratoire LIG, Université Grenoble Alpes
[3*] Computer Science, University of Neuchatel
[*] Corresponding author's email: Jacques.Savoy@unine.ch

**ABSTRACT**

This study explores the vocabulary richness of oral political communication. A model explaining the lexicon growth is proposed by subdividing the whole vocabulary into terms generated by general and specialized glossaries. The richness of vocabulary can be evaluated when based on a partition which is specific for each speaker. This study based on a corpus of candidates to the US presidential elections, demonstrates that Trump's language has the poorest vocabulary when compared to the others.

## 1 Introduction

In stylometry and quantitative linguistics, several studies have proposed different models to describe the lexical richness of a text. To achieve this, the most well-known approach is to explore the association between the number of word-tokens (or text length) and the number of word-types (or the vocabulary size). This relationship called Type-Token Ratio (or TTR) appears to be suitable for measuring approximatively the lexical sophistication or the richness of vocabulary. Such a feature can be employed to portray the general stylistic attribute of one work, to detect the style of a given author or to distinguish between various text genres (Kubát & Milička, 2013). As each writer has a distinct style, such a measure has been proposed as a means for authorship attribution (Hoover, 2003; Savoy, 2020). The vocabulary size and its relationship with the text coverage, vocabulary diversity and whether the discourse is comprehensible are also of prime interest for foreign language learners (Fengxiang, 2013; Hanušková et al., 2025). In computer science, this association can be employed to predict the memory space needed to manipulate textual data as, for example, when generating inverted files (Manning et al., 2008).

Besides these applications, the study of vocabulary richness may be used to observe rhetorical choice made by speakers in order to convince an audience. In this perspective, an orator might insist and repeat frequently the same essential arguments or expressions. During such text sequences, the speaker mainly reuses the same words or formulations and thus choses to reduce the vocabulary variety. On the other hand, a speaker might prefer to cover various topics or give details about his proposed solutions. This approach generates new terminology, producing a higher rate of new types (Hubert & Labbé, 1988**b**) and thus increases the vocabulary richness.

The main objective of this paper is to analyze the richness of the vocabulary in the transcripts of US political debates. During presidential campaign, in one to three TV debates, the two main presidential candidates reveal their solutions and provide answers to questions. Based on the transcripts of these debates, can we observe vocabulary size differences between candidates during the same election? Can we detect whether a candidate modifies his/her vocabulary richness during a debate, opting, for example, for simpler formulations? Over the years, does the vocabulary decrease leading to a simplification of political communication (Benoit et al., 2019)? Is Trump's vocabulary clearly poorer than the others? To what extent?

The rest of this paper is organized as follows. Section 2 presents an overview of related topics while Section 3 describes the corpus employed in our experiments. Section 4 exposes the proposed model to explain and measure the vocabulary richness. Using this approach, Section 5 analyzes the last three presidential elections with a particular focus on Trump. Section 6 compares the vocabulary richness of Trump's language with debaters prior to 2016. A conclusion reports the main findings of this study and the questions emerging from it.

## 2 Related Work

Counting words in a text is not straightforward and the definition of a word may vary. Some clarification is required. In the sentence "*I saw a man with a saw*", we count seven word-tokens but only five word-types (namely *I*, *saw*, *a*, *man*, *with*)[1]. Thus, a word-token (or simply token) is defined as an instance of a type. Measuring a text length means counting the number of tokens.

Different studies have proposed a model to measure the diversity and richness of the vocabulary occurring in a text. The most frequent model is based on the relationship between the number of types (or the vocabulary size) and the number of tokens (or Type-Token Ratio TTR) (Wimmer & Altmann, 1999; Baayen, 2008; Popescu et al., 2009; Mitchell, 2015; Hart, 2020). To compute this value, one can divide the vocabulary size (or the number of distinct types denoted by V) by the text length (number of tokens indicated by N) and we obtain TTR = V / N.

This approach corresponds to a global stylistic measure varying from ε (a small positive value) and 1.0. High values indicate the presence of a rich vocabulary showing that the underlying text reveals many various topics or that the author tends to present a theme with different formulations.

[1] Using the same example, one can find six lemmas (or dictionary entries) (namely I, see, a, man, with, saw).

One can cite another simple model proposed by Honoré (1979) who suggests a constant (denoted H) can be obtained when applying Equation 1. In this formulation, N indicates the text length, V the vocabulary size, and $V_1$ indicates the number of types appearing once (also called *hapax legomena*). In this case, $V_1$ is consider to reflect the vocabulary diversity of a text. A high $V_1$ value results in a high H score and signals a high vocabulary diversity.

$$H = 100 \cdot \frac{\log(N)}{1 - \left(V_1 / V\right)} \quad (1)$$

However, the previous mentioned models have a major drawback. When working with texts of different lengths, these measures are not really constant and thus do not correspond to a distinctive and stable stylistic feature. Moreover, they vary more than expected by pure chance (Baayen, 2001; 2008). This underlying instability is related to the fact that word frequency-based statistics correspond to the LNRE (Large Number of Rare Events) phenomenon (Baayen, 2008) in which a large number of types have very low probabilities of occurrence, which decrease while *N* grows.

To avoid this problem, a better computation is suggested by (Covington & McFall, 2010) or (Popescu et al., 2009). Instead of computing a single measure for a text, these authors suggest taking the moving average of TTR called MATTR (Kubát & Milička, 2013). Based on written speeches, the MATTR varies from one president to the next. For example, the highest score is achieved by Washington (41.2%) and the lowest by Wilson (36.1%) (Savoy, 2017).

As other possible measurements, one can consider the percentage of big words (BW) defined as words composed of six letters or more (at least for the English language) (Hart, 2020). One can observe that some terms are easier to understand than others as, for example, between "cars" and "automobiles" or "ads" and "advertisements". Such a relationship between complexity and word length is clearly established:

> "One finding of cognitive science is that words have the most powerful effect on our minds when they are simple. The technical term is basic level. Basic-level words tend to be short. … Basic-level words are easily remembered; those messages will be best recalled that use basic-level language." (Lakoff & Wehling, 2012, p. 41)

When a speech presents a high percentage of big words, its vocabulary is considered to be sophisticated and viewed as hard to follow and understand by the audience. For example, Fillmore (1850–1852) is the US president with the highest BW percentage (37.3%) compared to Clinton (25.1%) (Savoy, 2017).

As a fourth stylistic indicator, the lexical density (denoted LD) can be applied to reveal the capacity of a text to be informative (Biber et al., 2002), (Hewings et al., 2005). The formulation is shown in Equation 2 where the variable *N(t)* indicates the number of tokens of a text *t* (or its length), *function word(t)* the number of function words in *t*, *lexical word(t)* the number of lexical words in *t*. This latter set is composed of nouns, names, adjectives, verbs, and adverbs. On the other hand, function words regroup all other grammatical categories, namely determiners (e.g., the, this), pronouns (e.g., you, us), prepositions (e.g., to, in), conjunctions (e.g., and, or), modal verbs and auxiliary verb forms (e.g., has, would, can). As depicted in Equation 2, this LD value is given as a percentage over the text length.

$$\text{LD}(t) = \frac{lexical\ word(t)}{N(t)} = 1 - \left(\frac{function\ word(t)}{N(t)}\right) \quad (2)$$

A relatively high LD percentage indicates a more complex text, containing more information. Over all presidencies, the LD values varies from 50.3% (Eisenhower) who focusses more on topical forms and expressions to a minimal value of 41.5% (Wilson) (Savoy, 2017).

To the best of our knowledge, previous studies do not provide a clear and operational measure to define the vocabulary richness and to verify whether two values are statistically significantly different. Moreover, the distinction is not fully clear between diversity (or lexical variety), richness, sophistication (percentage of BW), and language density (LD). In addition, the findings of these previous studies are based on written production that, from a linguistic point of view, differs from oral one or web-based communication (Yule, 2020). As for example, one can expect more personal pronouns, verbs and adverbs and less complex sentences in an oral mode.

# 3 Corpus

To ensure our conclusions are founded on a strong base, a set of important considerations must be respected. Firstly, the individual making the speech must have written the text by themselves, rather than by a ghostwriter or a team of staffers. Therefore, selecting the oral communication mode is an evident and pertinent choice

Secondly, to compare meaningfully the speech of one person with another, the subject matter must cover similar topics and be generated under similar conditions. The set of presidential debates satisfies this constraint well. Over the last 60 years, the main candidates have been unwittingly participating in this sort of laboratory experiment: the presidential debates always follow the same format with equal speaking time, similar setting, and the same topics.

Thirdly, the collection of transcripts must be processed using to the same norms and respecting rigorous standards (Hockey & Martin, 1988). In this manner, each word can be clearly identified. Thus, the difference between the words starting with an uppercase letter or not can be considered as the same (e.g., the capitalized *The* or *the*). However, when the entire word occurs in uppercase, it could be considered as the same type as its corresponding lowercase type. Such matching is not always correct. For example, the token *US* corresponds to an abbreviation (*United States*) which is distinct from the pronoun *us*. Such distinctions are kept in this study (e.g., writing “U.S.”). Numbers have the same standardized form (e.g., 1,000 is represented as 1000) and the dollar sign is replaced by the term ‘dollars’.

As shown in Table 1, this corpus, containing 68 transcripts from 34 debates by 19 nominees in the US presidential election from 1960 to 2024, was downloaded from the web site *www.presidency.ucsb.edu*. During this time period, no presidential debates took place from 1964 to 1972. Therefore, this corpus contains close to the full set of debates broadcasted on different TV networks and cable news channels.

**Table 1.** Statistics about our corpus with the number of debates and the transcript length

| | Year | Debate | Length |
|---|---|---|---|
| Nixon | 1960 | 3 | 16,014 |
| Kennedy | 1960 | 3 | 14,828 |
| Ford | 1976 | 3 | 15,883 |
| Carter | 1976, 1980 | 3, 2 | 25,719 |
| Reagan | 1980 | 2 | 11,832 |
| Dukakis | 1988 | 2 | 16,132 |
| H. Bush | 1988, 1992 | 2, 3 | 30,663 |
| B. Clinton | 1992, 1996 | 3, 2 | 52,028 |
| Dole | 1996 | 2 | 18,263 |
| Gore | 2000 | 3 | 23,061 |
| G. Bush | 2000, 2004 | 3, 3 | 47,717 |
| Kerry | 2004 | 3 | 26,320 |
| McCain | 2008 | 3 | 23,598 |
| Obama | 2008, 2012 | 3, 3 | 51,077 |
| Romney | 2012 | 3 | 27,464 |
| H. Clinton | 2016 | 3 | 22,915 |
| Trump | 2016, 2020, 2024 | 3, 2, 2 | 64,958 |
| Biden | 2020, 2024 | 2, 1 | 24,652 |
| Harris | 2024 | 1 | 6,834 |

From this corpus, Trump appears having the largest number of texts (7) and tokens (64,958), followed by B. Clinton (5 texts and 52,028 tokens), and Obama (6 and 51,077). With the smallest number of tokens, Harris (one debate), is followed by Reagan (11,832) and Kennedy (14,828).

# 4 Vocabulary Richness

With texts of different lengths, the vocabulary is not directly comparable, except in certain cases. For example, during the first debate in 2016, Trump was more verbose than H. Clinton (8,402 tokens vs. 6,328), so one would expect Trump to employ more different types. However, the opposite was true (1,273 vs. 1,398 for H. Clinton). Another example: during the first 2024 debate, Trump's total vocabulary was 1,184 types (for 8,064 tokens spoken), fewer than the 1,193 employed by Biden for significantly shorter interventions (6,729 tokens). Likewise, during the second debate in 2024 (Harris vs. Trump), Harris used more types than Trump (1,238 vs. 1,229) even though she spoke much less than he did (5,676 vs. 8,081 tokens). It is therefore certain that H. Clinton (in 2016), Biden and Harris (in 2024) employed a richer vocabulary than Trump during these three debates.

In other cases, some computations are required. Consider two texts, namely A and B. Let Na and Nb be the lengths of the two texts (in tokens) and Va and Vb the number of different word types (or vocabulary).

If Na = Nb, the vocabulary of A will be considered richer if Va > Vb, or vice versa. However, as mentioned above, comparison is sometimes possible with texts of unequal lengths. Apart from these specific cases, how can one compare Va and Vb?

The size of the vocabulary V is a function of N (length of the text), but this function does not appear to follow any simple law, as suggested by the vocabulary growth profiles presented later.

The simplest approach could be the urn model. Let Na > Nb. One draws from A a sample of N'a = Nb tokens and count the V'a types contained in this sample. If V'a > Vb, it can deduce that A has a richer vocabulary than B, or vice versa. Naturally, the procedure will be lengthy since the operation must be repeated a large number of times and the average of the observations must be computed.

Muller (1977), Ule (1985) and Müller (2002) proposed formulae which simulate the result of these *exhaustive* polls (the token drawn is not put back into the urn).

The V types, in the whole work A, are graded in order of frequency into *I* frequency bins. Let Vi denotes the number of types occurring *i* times. N'k is the number of tokens up to the *k*th excerpt, for *i* = 1, 2, …, I. Following Muller (1977), the number of different types expected in this sample is given by the following equation.

(3) $V'(u) = V - \sum_{i=1}^{I} V_i \cdot Q_i(u)$ with $u = N'_k / N$, and $Q_i(u) = (1-u)^i$

Contrary to what Muller thought — calling this formula “binomial” — it yields the result of a large number of exhaustive polls (hypergeometric distribution) (Baayen, 2001) corresponding exactly to the case considered above (Hubert & Labbé, 1988a).

What results does Muller’s model gives when applied to debates in the US presidential elections?

For example, take Trump's and Biden’s speeches during the first debate in 2024. Since Trump's text is the longer one, it will be the subject of the computation. Boundaries are placed every 200 tokens, and we count Vk the number of types that have appeared from the beginning of the text up to the *k*th bound. Then we compute the expected (or theoretical) number of types at this same point k (V'k) using Equation 3.

In Figure 1, the bold line represents these theoretical values, and the dotted line reflects the number of types actually observed at the same intervals.

The observed values do not increase steadily. The curve exhibits fairly distinct undulations, and, most importantly, the theoretical number of types, calculated with Equation 3, is significantly higher than the observed ones. With rare exceptions, this is always the case with natural texts. For example, this is true for all debates in all US presidential debates since 1960. This bias means that, with this method, the richness of the vocabulary in the longest text will be overestimated compared to that of the shortest. In Figure 1, in the middle of the diagram, this overestimation is greater than 10%, which is an important discrepancy.

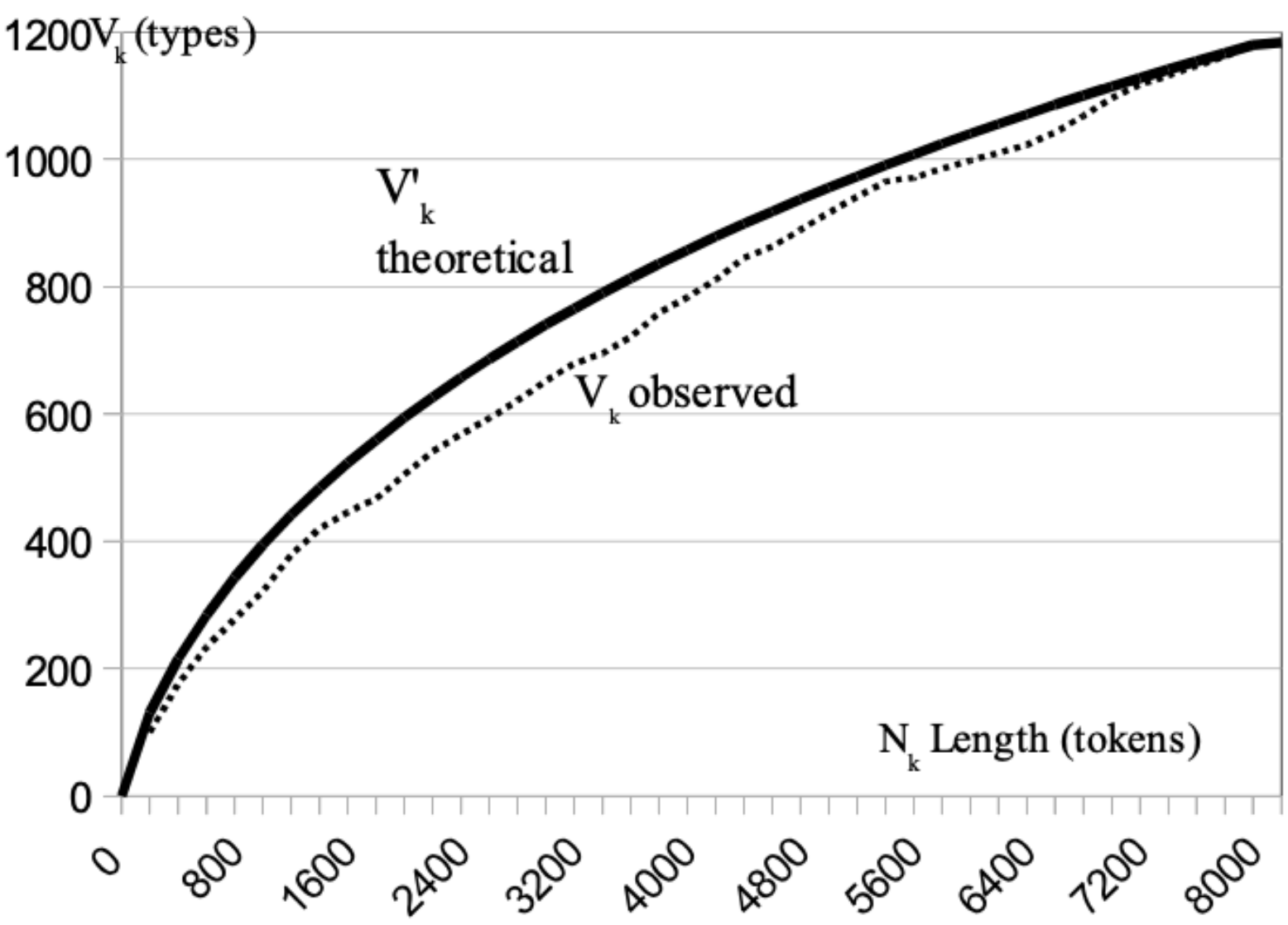

**Figure 1**. Vocabulary growth as a function of text length (in 200-token increments) in Trump's speeches compared to Biden's (first debate of 2024). Theoretical values (V'k) computed using the Muller model and observed values (Vk)

This bias stems from the fact that randomly drawing tokens from an urn destroys the internal logic of the text and no longer respects the syntax. Furthermore, the presence of certain types depends on the topic being addressed, and their probability of occurring is not equal throughout the text. For example, names of countries appear when discussing foreign policy, some content words are linked with specific topics (Kilgarriff,2001; Smadja, 1993; Church & Gale, 1999; Gerlach & Altmann, 2013). Since Muller (1977), this phenomenon has been referred to as the *vocabulary specialization*. We propose a model that adapts Muller's equation to account for this specialization.

## 4.1 The Vocabulary Partition Model

The emergence of new word types in a discourse, as the speech is given, can be described as a flow originating from two reservoirs (Hubert & Labbé, 1988b; 1997). This model has already been applied to several cases (e.g., Labbé & Labbé 2014). The current paper, besides presenting the calculation in a concise way, with simple examples, defines a procedure to compare a large number of speakers and texts, and to attach a level of confidence to the results.

The occurence of new word types corresponds to the following two sources. Firstly, there is general vocabulary, composed of types used regardless of the topic (grammatical words, verbs, nouns, and very common adjectives). As the text lengthens, the emergence of these types follows a hypergeometric pattern, comparable to the random selection described above:

$$\text{(4)} \qquad General\ vocabulary\ up\ to\ kth = (1-p)\cdot\left(V - \sum_{i=1}^{I} V_i \cdot Q_i(u)\right)$$

Here Formula 3 can be recognised but it is weighted by a parameter $p$ measuring the specialization of the vocabulary. The weight of the general vocabulary is therefore: 1-$p$. For example, $p$ = 0.270 in the Trump's speech facing Biden during the first 2024 debate (the calculation of $p$ is presented below). The rise of this general vocabulary ($GV_k$) follows the same pattern as the bold curve in Figure 1 above, but at a lower level (Figure 2).

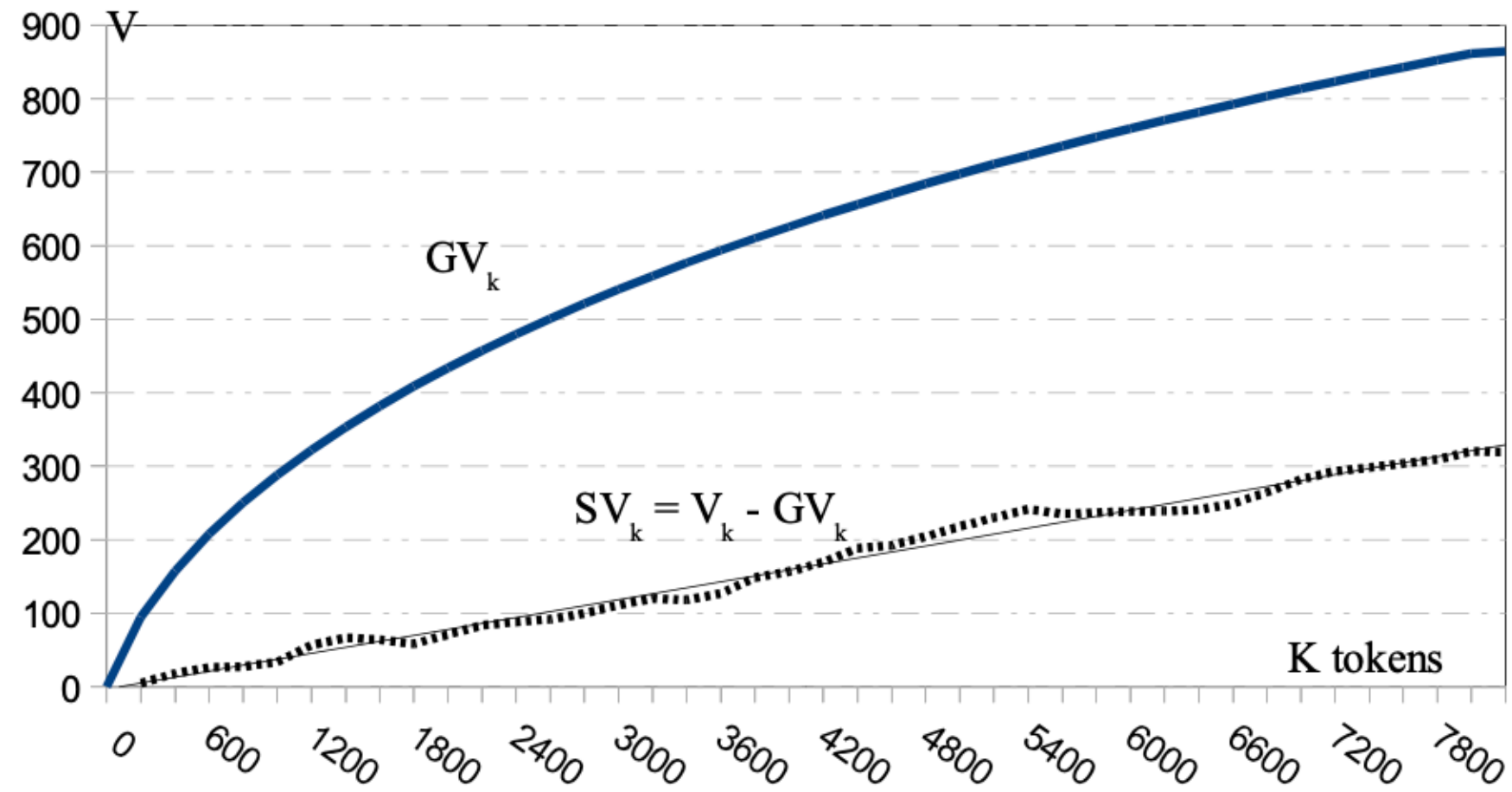

**Figure 2**. Increase in the two vocabularies (general and specialised) as a function of text length in Trump's speech (first 2024 debate). Theoretical values for the general vocabulary ($GV_k$) and observed values for the specialised one ($SV_k$).

Secondly, the specialised vocabulary (SV). The number of specialised word types, at the *k*th bounday, is: $SV_k = V_k - GV_k$ (dotted line). As shown in Figure 2, the growth trend of this vocabulary (plain thin line) is linear and its slope depends only on *p*, subject to a few minor fluctuations, which show that Trump's tendency to specialize his vocabulary is not completely stable.

The theoretical volume of this specialised vocabulary, at the *k*th boundary, is given by p • V • u. In other words, the partitionning model calculates the weigth of the two vocabularies but is unable to point out which words types come from the general or from the specialised one.

The formula applied to compute $V'_k$ is the sum of these two vocabularies, namely:

$$V'(u) = p \cdot V \cdot u + (1-p) \cdot \left(V - \sum_{i=1}^{I} V_i \cdot (1-u)^i\right) \quad (5)$$

The partition parameter that gives the relative weight of these two reservoirs is computed using the well-known least squares method in order to minimize the distance between the observed $V_k$ and theoretical $V'_k$ as depicted in Formula 6.

$$p = \frac{\sum_{k=1}^{K}\left((u_k-1)\cdot V+\sum_{i=1}^{I} V_i \cdot Q_i(u_k)\right)\cdot\left(V'(u_k)-V+\sum_{i=1}^{I} V_i \cdot Q_i(u_k)\right)}{\sum_{k=1}^{K}\left((u_k-1)\cdot V+\sum_{i=1}^{I} V_i \cdot Q_i(u_k)\right)^2} \quad (6)$$

For example, in Trump's speech during the first 2024 debate (see Figure 1), $p = 0.270$. In other words, of the 1,184 types uttered by Trump during this debate, 320 belonged to specialized vocabulary and 864 to general vocabulary. As displayed in Figure 3, this model thus provides a precise fit to the observed curve.

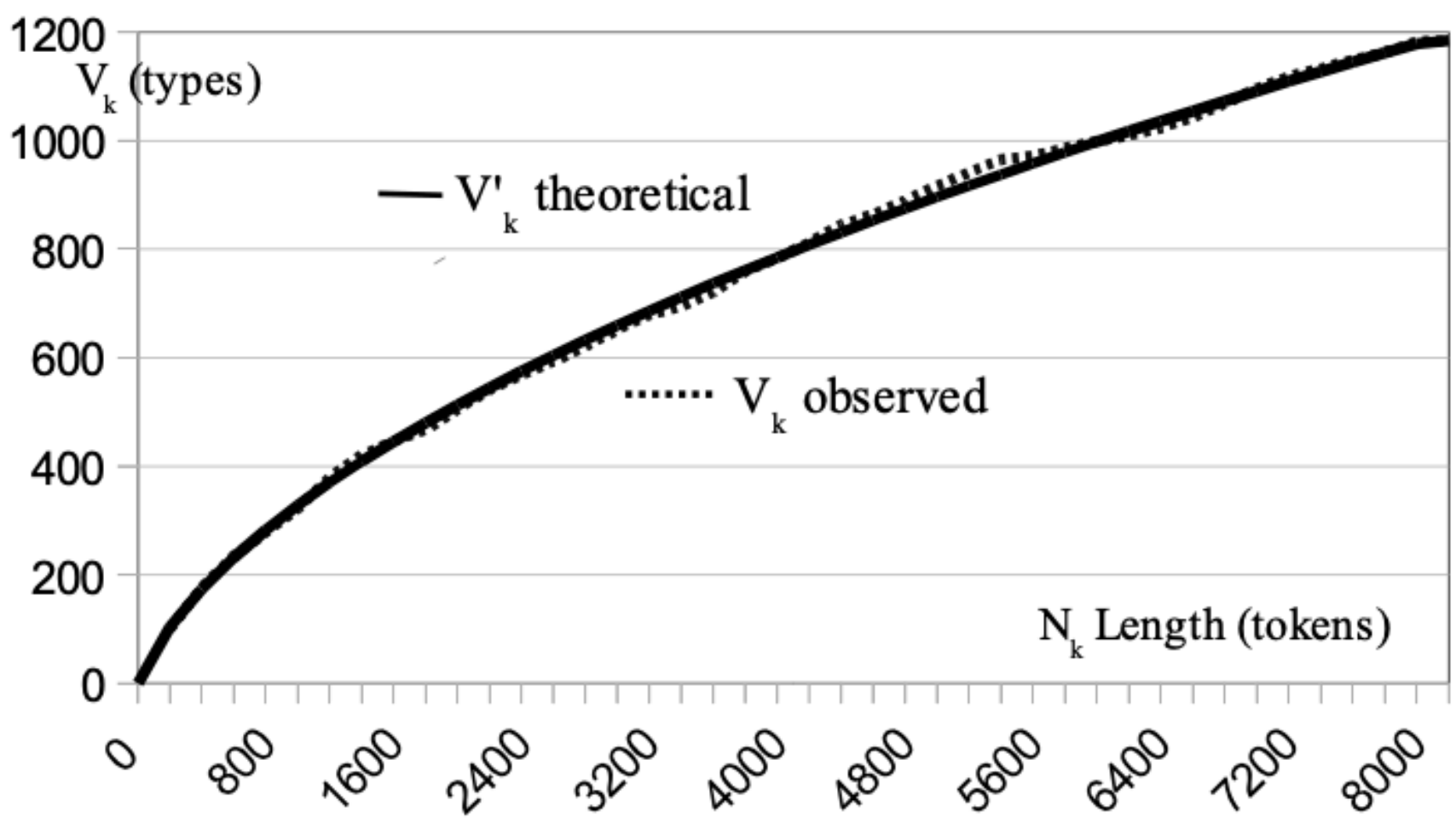

**Figure 3**. Increase in vocabulary as a function of text length in Trump's speech (first 2024 debate). Theoretical values ($V'_k$) computed using the vocabulary partitioning model and observed values ($V_k$).

The observed values still exhibit the same small anomalies, but this time they are evenly distributed around the values computed using the partition model. For the *K* bounds, the sum of the differences between the observed values ($V_k$) and the computed values $V'_k$) is near zero (since V' are positive decimals and V are integers).

$$\sum_{i=1}^{K}(V_i - V'_k) \approx 0 \qquad (7)$$

The partitioning model thus offers an unbiased representation of the phenomenon, allowing us to compare texts with different lengths. It is now possible to provide an answer to the question "Who has the richest vocabulary", i.e., how many types would the longer text (Trump's one) have contained if it had been the same length as the shorter one (Biden's). Figure 4 displays the application for this 2024 debate.

If the number of types in the shorter text falls outside this interval, one can accept the hypothesis that there is a significant difference in vocabulary richness between the two texts or two speakers. This acceptation is stronger when the observed vocabulary in the shorter text is far from the interval limits.

Biden uttered 6,729 tokens (compared to 8,064 for Trump) and employed a vocabulary of 1,193 distinct types. The partitioning model indicates that, if Trump's speech has been the same length as Biden's, it would have contained 1,062 different types, 11% fewer than the incumbent Democrat president, which is significant.

The basic premise of this reasoning is that vocabulary increases uniformly with text length. However, in the figures, the observed values — the dotted lines — do not have the uniform appearance of the theoretical values. Therefore, two questions arise. Firstly, can we consider these to be really random fluctuations that can be observed in many natural phenomena? Secondly, do the differences between the two texts exceed these natural fluctuations? If so, we can conclude, with a low risk of error, that there is a significant difference between the vocabulary richness of the two speeches.

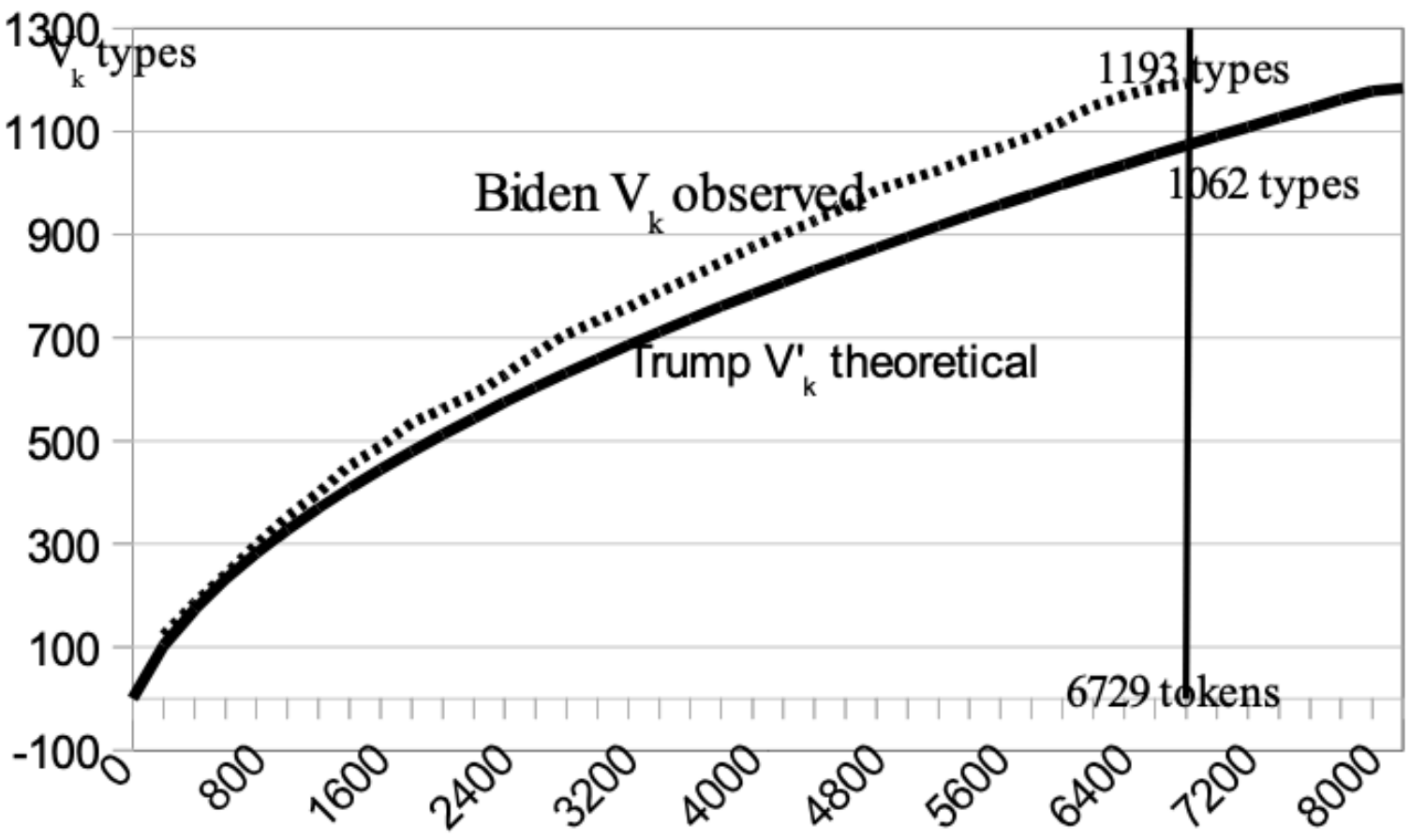

**Figure 4**. Comparison of vocabulary growth in speeches by Biden and Trump (relative to the length of Biden's speech)

To answer these two questions, we associate with the theoretical values, standard deviations, denoted σ, which are the square root of variance and can be computed according to Equation 8 (Labbé & Labbé, 2014).

$$(8) \qquad Var[V'(u)\,] \;=\; \sum_{i=1}^{K} Q_i(u) \cdot (1 - Q_i(u))$$

This allows us to define a "normal" or "expected fluctuation interval" along the curve, given an α risk of error.

For example, for the first debate Biden vs. Trump, with α = 5%, this range surrounding the expected value (1,062 types) is defined as:

Lower bound: 1,062 - 7.4 · 1.96 = 1,048 types

Upper bound: 1,062 + 7.4 · 1.96 = 1,077 types

Since the 1,193 types uttered by Biden are far above this upper bound, one can conclude, with a very little risk of error, that, during this debate, Trump employed a poorer vocabulary than the incumbent democrat.

This interval is dependent on the length, as shown in Figure 5. To make this correlation clear, the trend of the expansion is displayed by the horizontal axis.

This characteristic has an important consequence for the method presented below: the greater the differences in length between the two texts being compared, the larger the range of uncertainty will be.

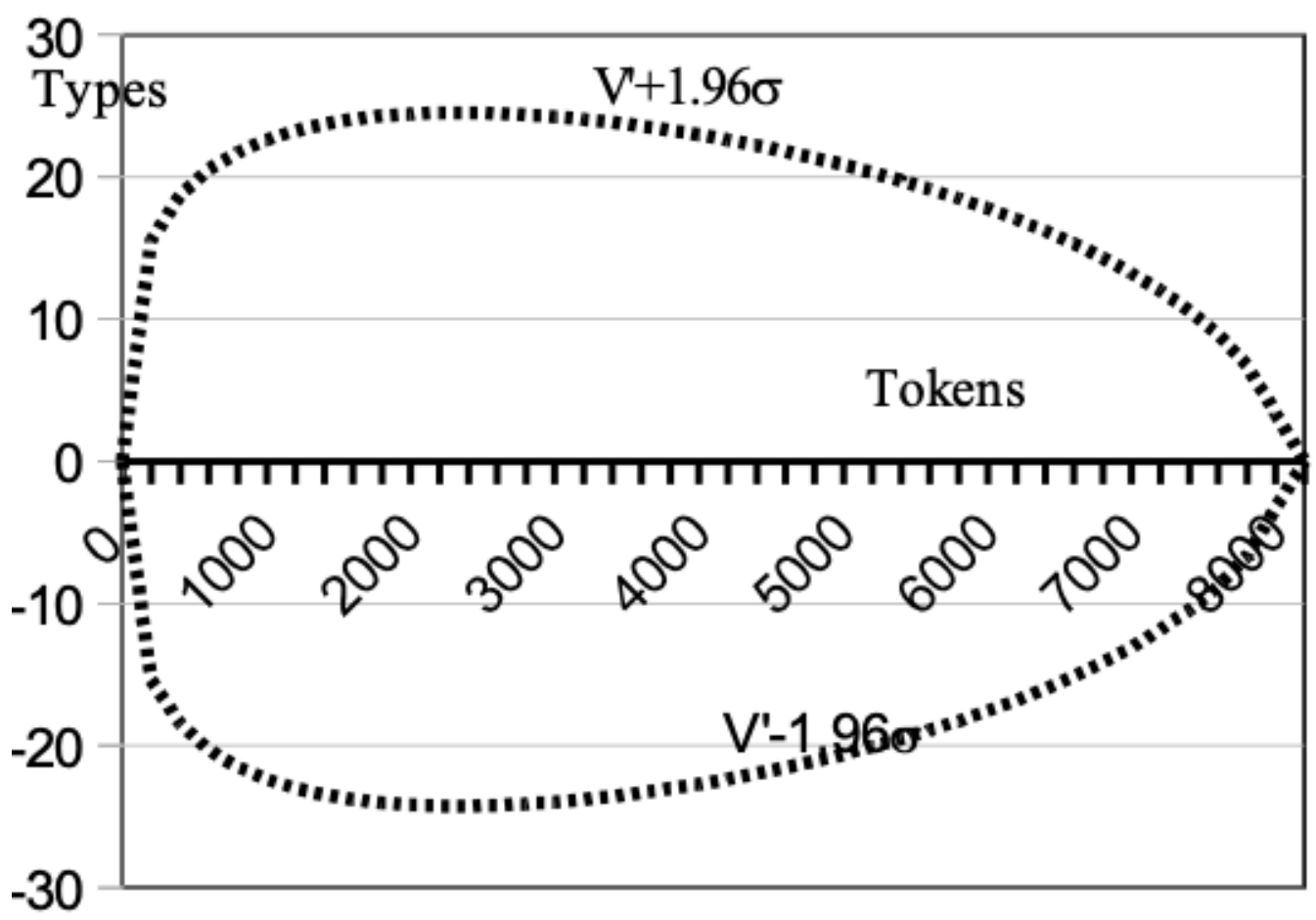

**Figure 5**. Evolution of the normal fluctuation interval as a function of the length of N' (with α = 5%) (Trump, first debate, 2024)

With α = 5%, for a single text, if 95% of the observed values fall within this interval, we cannot reject the hypothesis of an uniform increase in vocabulary throughout the text. In addition, values outside the interval indicate thematic or stylistic breaks, and we can precisely locate these breaks within the text. But this is beyond the scope of the present discussion.

## 5 Trump in the Debates from 2016 to 2024

Firstly, our emphasis is on the last three editions (2016, 2020 and 2024) in which Trump appears as shown in Table 2. During these last three elections, Harris delivered the shortest speech: 5,676 tokens as shown in the second column of Table 2 (labelled N). The diversity of the other participants' vocabulary (V) is measured by this length as depicted in the third column.

Therefore, it is proposed to compute V' along this dimension in the fourth column of Table 2. In this table, the speeches are ranked in ascending order (from poorest to richest). A standard deviation (σ) is associated with these *theoretical* values (V'). This allows us to assign each text a diversity index within a certain range of fluctuation (5% and 1%).

The debaters were ostensibly allotted roughly the same speaking time, but they did not utter the same number of tokens. In all three elections, Trump consistently managed to employ the most tokens, except for the third debate in 2016 (which focused on social issues: healthcare, education, immigration, and public safety), during which H. Clinton spoke slightly more.

Two main conclusions can be drawn.

Firstly, a significant shift can be observed between Trump and his competitors. Compared to Trump's speech during the first 2024 debate, H. Clinton, in 2016 — on the same topics (domestic economy) — used 33% more words, or a third more. This is a considerable difference! Had she memorized her speech? Given the structure of the debates (numerous questions and brief answers), this seems unlikely, and it's more likely that her training and experience speaking to demanding audiences (Congress, committees, the media, etc.) influenced her style, which was quite different from everyday conversation.

**Table 2.** The richness of vocabulary in the debates of 2016, 2020, and 2024

| | N | V | $V'_{5676}$ | σ | P |
|---|---|---|---|---|---|
| H. ClintonA | 6,328 | 1,398 | 1,287 | 6.6 | 443 |
| H. ClintonB | 6,286 | 1,306 | 1,221 | 7.6 | 233 |
| H. ClintonC | 6,999 | 1,434 | 1,255 | 10.1 | 328 |
| Trump2016A | 8,402 | 1,273 | 1,001 | 10.4 | 285 |
| Trump2016B | 7,349 | 1,154 | 1,017 | 8.80 | 255 |
| Trump2016C | 6,690 | 1,076 | 972 | 6.60 | 328 |
| Biden2020A | 6,630 | 1,214 | 1,100 | 7.67 | 261 |
| Biden2020B | 6,975 | 1,331 | 1,152 | 8.10 | 403 |
| Trump2020A | 7,025 | 1,120 | 982 | 7.50 | 325 |
| Trump2020B | 7,074 | 1,157 | 1,019 | 8.61 | 20 |
| Biden2024 | 6,729 | 1,193 | 1,079 | 7.92 | 216 |
| Harris | **5,676** | 1,238 | 1,238 | 0 | 152 |
| Trump2024A | 8,064 | 1,184 | **965** | 9.53 | 270 |
| Trump2024B | 8,081 | 1,229 | 1,004 | 10.47 | 202 |

Secondly, three categories emerge.

- Low diversity: all of Trump's speeches with slight variations over time and depending on the topics, with those variations being not significant at the 1% level.

**Table 3.** The richness of the vocabulary in the debates of 2016, 2020, and 2024 with the lower and upper limits (1%, 5%) (texts corresponding to the length of Harris' interventions)

| | – 1% | -5% | $V'_{5676}$ | + 5% | +1% | Index |
|---|---|---|---|---|---|---|
| Trump2024 | 941 | 946 | 965 | 984 | 989 | 100 |
| Trump2016C | 955 | 959 | 972 | 985 | 989 | 101 |
| Trump2020A | 963 | 967 | 982 | 997 | 1,001 | 102 |
| Trump2016A | 974 | 981 | 1,001 | 1,021 | 1,028 | 104 |
| Trump2024B | 977 | 983 | 1,004 | 1,024 | 1,031 | 104 |
| Trump2016B | 994 | 1,000 | 1,017 | 1,034 | 1,040 | 105 |
| Trump2020B | 997 | 1,002 | 1,019 | 1,036 | 1,041 | 106 |
| Biden2024 | 1,059 | 1,063 | 1,079 | 1,094 | 1,099 | 112 |
| Biden2020A | 1,080 | 1,085 | 1,100 | 1,115 | 1,120 | 114 |
| Biden2020B | 1,131 | 1,136 | 1,152 | 1,168 | 1,173 | 119 |
| ClintonHB | 1,202 | 1,206 | 1,221 | 1,236 | 1,240 | 127 |
| Harris | 1,238 | 1,238 | 1,238 | 1,238 | 1,238 | 128 |
| ClintonHC | 1,229 | 1,235 | 1,255 | 1,275 | 1,281 | 130 |
| ClintonHA | 1,270 | 1,274 | 1,287 | 1,300 | 1,304 | 133 |

- Average diversity: Biden depicts a slightly larger range. It is remarkable to note that, during the first debate of 2024, despite obvious speech difficulties, he performed only slightly worse than during the first debate of 2020, at least in terms of vocabulary diversity.

- High diversity: H. Clinton and Harris.

Considering all Trump’s speeches from 2016 to 2024, the overlap of the intervals is perfect, yielding a single interval ranging from 941 to 1,041 types for 5,676 tokens (see Table 3). He therefore demonstrated great stylistic stability. This upper limit is significantly lower than the smallest value in this same interval for Biden. Therefore, there is less than a 1% chance of being wrong in stating that the diversity of vocabulary used by Trump is *stable* and *significantly lowe*r than that of his three opponents.

Is this also the case for the other debates since 1960?

# 6 Comparison with Debates Prior to 2016

To compare with televised debates from 1960 to 2012, it is proposed to consider only Trump's “richer” speech: the second debate of 2020 containing 1,157 types for 7,074 tokens. The question then becomes as follows: since 1960, have there been any debate with a vocabulary diversity equal to or less than this Trump’s “better” performance?

The procedure will differ depending on whether the text being compared is shorter or longer than the Trump2020B (7,074 tokens).

- For shorter texts, V' is computed using the length of Trump2020B as for example with Kennedy1960A with 4,695 tokens as displayed in Table 4. The following questions are examined: if in October 2020, Trump had only used 4,625 tokens, how many types (V') would he have uttered? What are the confidence intervals for this theoretical value? Does Kennedy's vocabulary (953 types) fall outside these intervals?

- Longer texts are scaled to the length of Trump2020B (7,074 tokens), with the theoretical value associated with two confidence intervals.

## 6.1 With Shorter Texts

In Table 4, the texts are ranked by increasing diversity relative to Trump's speech during the first debate of 2020.

The first line shows that, if Trump had uttered only 4,692 tokens (N) in October 2020, he would have used 912 types ($V'_N$) whereas Nixon's speech during the first debate contained only 874, or 4% fewer as shown in the last column of Table 4. Since the lower bound of the confidence intervals, 885 types, is significantly higher than Nixon's performance, we can conclude that, during his best performance (first debate, 2020), Trump used a slightly richer vocabulary than Nixon (during this first confrontation with Kennedy). We also note that, for Nixon's third performance (second line), the upper limit of the interval and the vocabulary observed in Nixon are equal (935 types). Therefore, we cannot reject the hypothesis - that the two men had the same vocabulary diversity - with less than a 1% probability of being wrong.

**Table 4.** Comparisons with Trump's first debate 2020. Texts are ranked by increasing diversity

| | N | V | $V'_{(N)}$ | σ | -1% | -5% | +5% | +1% | (V-V')/V |
|---|---|---|---|---|---|---|---|---|---|
| Nixon1960A | 4,692 | 874 | 912 | 10.7 | 885 | 891 | 933 | 939 | -4% |
| Nixon1960C | 4,652 | 935 | 908 | 10.7 | 881 | 887 | 929 | 935 | 3% |
| Ford1976A | 4,580 | 935 | 900 | 10.8 | 872 | 879 | 921 | 928 | 4% |
| Nixon1960B | 4,753 | 977 | 919 | 10.6 | 892 | 898 | 940 | 946 | 6% |
| BushGH1992C | 4,960 | 1,027 | 942 | 10.2 | 916 | 922 | 962 | 968 | 8% |
| Kennedy1960B | 4,697 | 998 | 913 | 10.7 | 886 | 892 | 934 | 940 | 9% |
| BushGW2004A | 6,385 | 1,199 | 1,091 | 6.3 | 1,075 | 1,079 | 1,103 | 1,107 | 9% |
| Ford1976B | 4,668 | 1,005 | 910 | 10.7 | 883 | 889 | 931 | 937 | 9% |
| Obama2012C | 6,718 | 1,242 | 1,123 | 4.6 | 1,111 | 1,114 | 1,132 | 1,135 | 10% |
| ClintonW1992C | 4,897 | 1,047 | 935 | 10.3 | 909 | 915 | 955 | 961 | 11% |
| Dukakis1988B | 6,912 | 1,288 | 1,142 | 3.1 | 1,134 | 1,136 | 1,148 | 1,150 | 11% |
| BushGW2000C | 6,514 | 1,249 | 1,103 | 5.7 | 1,088 | 1,092 | 1,114 | 1,118 | 12% |
| BushGH1992A | 5,127 | 1,093 | 961 | 9.9 | 936 | 942 | 980 | 986 | 12% |
| Kennedy1960C | 3,859 | 940 | 814 | 11.8 | 784 | 791 | 837 | 844 | 13% |
| ClintonW1992B | 4,940 | 1,087 | 940 | 10.2 | 914 | 920 | 960 | 966 | 14% |
| Kennedy1960A | 4,695 | 953 | 814 | 11.8 | 784 | 791 | 837 | 844 | 15% |
| MacCain2008B | 6,499 | 1,291 | 1,102 | 5.8 | 1,087 | 1,091 | 1,113 | 1,117 | 15% |
| BushGW2004B | 6,682 | 1,313 | 1,120 | 4.8 | 1,108 | 1,111 | 1,129 | 1,132 | 15% |

For all the others, including Nixon's second speech, Trump's vocabulary is significantly less extensive. These differences should be viewed as important, as indicated in the last column of Table 4.

### 6.2 With Longer Texts

These texts determine the theoretical vocabulary volume if they had been the length of Trump's “better” speech. The texts are ranked by increasing diversity.

As depicted in the line of Table 5, for Romney, during the first 2012 debate against Obama, with α = 5%, we can reject the hypothesis of a vocabulary as limited as Trump's, but not at the 1% threshold: the lower bound is equal to Trump's better diversity (1,157 types).

All other candidates who spoke more than 7,074 tokens used a more diverse vocabulary than any of Trump's speeches. Except for Romney's first two debates, the differences are significant, ranging from +9% to +22%, as indicated in the last column of Table 5.

**Table 5.** Comparison with longer texts reduced to the length of the first debate
(Trump, first debate 2020: 7,074 tokens and 1,157 types)

| | N | V | $V'_{7074}$ | σ | -1% | -5% | +5% | +1% | (V'-1157)/1157 |
|---|---|---|---|---|---|---|---|---|---|
| Romney2012A | 7,824 | 1,247 | 1,172 | 6 | 1,157 | 1,160 | 1,184 | 1,187 | 1% |
| Romney2012B | 7,305 | 1,197 | 1,173 | 3.3 | 1,165 | 1,167 | 1,179 | 1,181 | 1% |
| Obama2012B | 7,558 | 1,316 | 1,260 | 5 | 1,247 | 1,250 | 1,270 | 1,273 | 9% |
| Obama2012A | 7,326 | 1,298 | 1,271 | 4.1 | 1,261 | 1,263 | 1,279 | 1,287 | 10% |
| Romney2012C | 8,094 | 1,378 | 1,273 | 8 | 1,253 | 1,257 | 1,289 | 1,293 | 10% |
| Dukakis1988A | 7,171 | 1,286 | 1,276 | 2.5 | 1,270 | 1,271 | 1,281 | 1,282 | 10% |
| Dole1996B | 7,244 | 1,294 | 1,278 | 3.7 | 1,269 | 1,271 | 1,285 | 1,284 | 10% |
| Kerry2004A | 7,438 | 1,327 | 1,290 | 5.2 | 1,277 | 1,280 | 1,300 | 1,298 | 11% |
| Dole1996A | 8,198 | 1,415 | 1,308 | 9 | 1,285 | 1,290 | 1,326 | 1,331 | 13% |
| BushG2000A | 7,292 | 1,342 | 1,318 | 3.9 | 1,308 | 1,310 | 1,326 | 1,328 | 14% |
| ClintonW1996A | 7,725 | 1,388 | 1,325 | 7 | 1,307 | 1,311 | 1,339 | 1,343 | 15% |
| BushG2000B | 7,580 | 1,396 | 1,327 | 5.1 | 1,314 | 1,317 | 1,337 | 1,340 | 15% |
| Obama2008B | 7,128 | 1,351 | 1,345 | 2.1 | 1,340 | 1,341 | 1,349 | 1,350 | 16% |
| Kerry2004C | 7,097 | 1,360 | 1,357 | 1.3 | 1,354 | 1,354 | 1,360 | 1,361 | 17% |
| Gore2000A | 7,369 | 1,393 | 1,358 | 4.5 | 1,346 | 1,349 | 1,367 | 1,370 | 17% |
| Obama2008A | 7,631 | 1,427 | 1,361 | 6.7 | 1,344 | 1,348 | 1,374 | 1,378 | 18% |
| Obama2008C | 7,250 | 1,401 | 1,378 | 3.5 | 1,369 | 1,371 | 1,385 | 1,387 | 19% |
| Kerry2004B | 7,513 | 1,468 | 1,411 | 5.5 | 1,397 | 1,400 | 1,422 | 1,419 | 22% |

# 7 Conclusion

This paper presents a model to measure the vocabulary richness of a text, speech, or collection of works. This approach is based on the split of the entire vocabulary into two distinct subsets, namely a general and a specialized one. The latter one tends to increase linearly with the text length. The growth of the general vocabulary composed of common words (regardless of the topic such as functional terms, nouns, verbs, adverbs, and adjectives) follows a more complex curve.

Based on this model, the size of the two subsets of the entire vocabulary can be estimated. In addition, when comparisons are required, the upper and lower limit of the vocabulary size can be accurately defined. This characteristic allows the experimenter to specify when a difference in vocabulary size can be viewed as statistically significant.

In our analysis of Trump's speeches during the debates in the presidential elections, one can conclude that they are barely superior to Nixon's first debate with Kennedy in 1960. But in the two subsequent debates, Nixon more or less corrected this flaw. Similarly, using the strictest criteria, it is impossible to distinguish Romney's shortest speech from Trump's better one.

Beyond this, Trump's vocabulary is certainly more limited than that of all the presidents who preceded him and almost all of their challengers. The differences are statistically significant. It is therefore clear that Trump does not speak like any other candidate to the US presidency.

This characteristic may stem from a communication strategy, as Trump claims to break with *Washington politicians* who, in his own words, "talk the talk but do nothing." However, it could also be a personality trait, since one does not speak like that for hours without revealing something about oneself.

Further analysis will be naturally needed to reveal the specific characteristics of Trump's style, particularly his preference for the most common verbs or the difference between written and oral communication modes. However, in this study based on oral mode, gesture, facial expressions, intonation, prosody, and volume of the speaker’s voice have not been studied.

The tool has other potential uses, including the detection of thematic or stylistic shifts within a work or a corpus.

## Acknowledgements

The authors want to thank Prof. Edouard Arnold (Trinity College – Dublin) for his comments and remarks on a first draft of this manuscript and the anonymous referees for their suggestions to improve the readiness of this article.

## References

**Baayen, H.R**. (2001). *Word Frequency Distribution*. Dordrecht : Kluwer.

**Baayen, H.R**. (2008). *Analysis Linguistic Data: A Practical Introduction to Statistics Using R*. Cambridge University Press: Cambridge.

**Benoit, K., Munger, K., & Spirling, A**. (2019). Measuring and explaining political sophistication through textual complexity. *American Journal of Political Science*, 63(2), pp. 491–508. doi.prg/10.1111/ajps.12423

**Biber, D., & Conrad, S**. (2009). *Register, Genre, and Style*. Cambridge: Cambridge University Press

**Covington, M.A., & McFall, J.D**. 2010. Cutting the Goridan knot: The moving-average type-token ratio (MATTR). *Journal of Quantitative Linguistics*, 17(2), pp. 94–100, doi.org/10.1080/09296171003643098

**Church, K., & Gale, W.** 1999. Inverse document frequency (idf): A measure of deviations from Poisson, in: *Natural Language Processing Using Very Large Corpora*. Springer, pp. 283–295. https://link.springer.com/chapter/10.1007/978-94-017-2390-9_18

**Fengxiang, F**. (2013). Text Length, vocabulary size and text coverage constancy. *Journal of Quantitative Linguistics*, 20(4), 288-300. doi.org/10.1080/09296174.2913.830550

**Gerlach, M., & Altmann, E.G.** (2013). Stochastic model for the vocabulary growth in natural languages. *Physical Review*, X(3), 021006. doi.org/10.1103/PhysRevX.3.021006

**Hanušková, M., Kubát, M., & Nogolová, M**. (2025). Lexical diversity of Czech L.2 texts at different proficiency levels. *Glottometrics*, 59, pp. 44–55. https://doi.org/10.53482/2025_59_427

**Hart, R.P**. (2020). *Trump and Us*. *What He Says and Why People Listen*. Cambridge: Cambridge University Press.

**Hewings, A., Painter, C., Polias, J., Dare, B., & Rhys, M.** (2005). Getting started: Describing the grammar of speech and writing. Milton Keynes: *Open University Press*

**Hockey S., & Martin J.** (1988). *OCP Users' Manual*. Oxford: Oxford University Computing Service.

**Honoré, A.** (1979). Some simple measures of richness of vocabulary. *Association of Literary & Linguistic Computing Bulletin*, 7(2), pp. 172–179.

**Hoover, D.L.** (2003). Another perspective on vocabulary richness. *Computers and the Humanities*, 37, 151–178. https://www.jstor.org/stable/30204890

**Hubert, P., & Labbé, D.** (1997). Vocabulary richness. *Lexicometrica*. n° 0, hiver 1997-98. https://www.researchgate.net/publication/237536078

**Hubert, P., & Labbé, D.** (1988a). Note sur l'approximation de la loi hypergéométrique par la formule de Muller. In Labbé, D., Serant, D. & Thoiron, P. (Eds), *Etudes sur la richesse et la structure lexicales*. Genève-Paris : Slatkine-Champion, 1988, pp. 77-91. https://www.researchgate.net/publication/281418955

**Hubert, P., & Labbé, D.** (1988b). A model of vocabulary partition. *Literary and Linguistic Computing*. 3(4), pp. 223-225. https://www.researchgate.net/publication/278817368

**Kilgarriff, A.** (2001). Comparing corpora. *International Journal of Corpus Linguistics*, 6(1), pp. 97–133.

**Kubát, M., & Milička, J.** (2013). Vocabulary Richness Measures in Genres. *Journal of Quantitative Linguistics*, 20(4), pp. 339–349. doi.org/10.1080/09296174.2013.830552

**Labbé, C., & Labbé, D.** (2014). Was Shakespeare's vocabulary the richest? In Née E., Daube J-M. Valette M., Fleury S. (Eds), *Proceedings of the 12th International Conference on Textual Data Statistical Analysis*. Paris: June 3-6 2014, pp. 323-336.

**Labbé, C., Labbé, D., & Hubert, P**. (2004). Automatic segmentation of texts and corpora. *Journal of Quantitative Linguistics*, 11(3), pp. 193–213. https://www.researchgate.net/publication/220469438

**Lakoff, G., & Wehling, E**. (2012). *The Little Blue Book: The Essential Guide to Thinking and Talking Democratic*. New York: Free Press.

**Manning, C.D., Raghavan, P., & Schütze, H.** (2008). *Introduction to Information Retrieval.* Cambridge: The Cambridge University Press.

**Mitchell, D**. (2015). Type-token models: A comparative study. *Journal of Quantitative Linguistics*, 22, pp. 1-21. doi.org/10.1080/09296174.2014.974456

**Muller, C.** (1977). *Principes et méthodes de statistique lexicale.* Paris: Hachette.

**Müller, D**. (2002). Computing the type token relation from the *a priori* distribution of types. *Journal of Quantitative Linguistics*, 9(3), pp. 193–214.

**Popescu, I.-I., Atlman, G., Grzybek, P., Jayaram, B. D., Köhler, R., Krupa, V., Macutek, J., Pustet, R., Uhlirova, L., & Vidya, M.N**. (2009). *Word Frequency Studies*. Berlin: De Gruyter Mouton.

**Savoy, J**. (2017). Analysis of the style and the rhetoric of the American presidents over two centuries. *Glottometrics*, 38, pp. 55–76. https://www.researchgate.net/publication/317304389_Analysis_of_the_Style_and_the_Rhetoric_of_the_American_Presidents_Over_Two_Centuries

**Savoy, J.** (2020). *Machine Learning Methods for Stylometry. Authorship Attribution and Author Profiling*. Cham: Springer.

**Smadja, F**. (1993). Retrieving collocations form text: Xtract. Computational Linguistics, 19(1), pp. 143-177. https://aclanthology.org/J93-1007/

**Ule, L**. (1985). The weird ways of vocabulary. *Literary and Linguistic Computing Journa*l. 6(1), pp. 24-28.

**Wimmer, G., & Altmann, G**. (1999). *Review Article: On vocabulary richness*. *Journal of Quantitative Linguistics*, 6(1), 1-9.

**Yule, G**. (2020). *The Study of Language*. Cambridge: Cambridge University Press.